# Fairness-Aware Interpretable Modeling (FAIM) for Trustworthy Machine Learning in Healthcare

Mingxuan Liu[1], Yilin Ning[1], Yuhe Ke[2], Yuqing Shang[1], Bibhas Chakraborty[1,3,4,5], Marcus Eng Hock Ong[3,6], Roger Vaughan[1,3], Nan Liu[1,3,7]*

[1] Centre for Quantitative Medicine, Duke-NUS Medical School, Singapore, Singapore

[2] Department of Anaesthesiology and Perioperative Medicine, Singapore General Hospital, Singapore

[3] Programme in Health Services and Systems Research, Duke-NUS Medical School, Singapore, Singapore

[4] Department of Statistics and Data Science, National University of Singapore, Singapore

[5] Department of Biostatistics and Bioinformatics, Duke University, Durham, NC, USA

[6] Department of Emergency Medicine, Singapore General Hospital, Singapore, Singapore

[7] Institute of Data Science, National University of Singapore, Singapore, Singapore

*Correspondence: Nan Liu, Centre for Quantitative Medicine, Duke-NUS Medical School, 8 College Road, Singapore 169857, Singapore

Email: liu.nan@duke-nus.edu.sg

## Abstract

### Background

The escalating integration of machine learning in high-stakes fields such as healthcare raises substantial concerns about model fairness and interpretability. Conventional bias-mitigation techniques often sacrifice model performance (e.g., accuracy), limiting their impact on aiding decision-making. This study introduces an interpretable bias mitigation framework, Fairness-Aware Interpretable Modeling (FAIM), aiming to balance model performance and fairness.

### Methods

We propose the FAIM framework, which prioritizes fairness in predictive models that are nearly optimal in performance. Initiating with a performance-optimized model yet without explicit fairness considerations (i.e., "fairness-unaware"), we construct a set of nearly-optimal models (i.e., Integral Rashomon Set) to select a fairer alternative (i.e., "fairness-aware"). To identify this fairness-aware model, we introduce a fairness ranking index (FRI) to comprehensively rank model fairness, which encompasses multiple fairness metrics including equalized odds, equal opportunity, and equality of balanced error rate. This index is complemented by an interactive graphical interface that facilitates the incorporation of clinical expertise to guide model selection with contextualized fairness. Moreover, we integrate SHapley Additive exPlanations (SHAP) within FAIM to compare variable contributions between the fairness-unaware and fairness-aware models, thereby uncovering model modifications made to address fairness. We demonstrated the value of FAIM by predicting hospital admission using two real-world databases, MIMIC-IV-ED and SGH-ED, to test its performance in bias mitigation, particularly for sex and race. We gauged model performance via AUC, sensitivity and specificity, and evaluated model fairness via the aforementioned fairness metrics.

**Results**

For both datasets,-the fairness-aware models yielded by FAIM automatically excluded race and reduced the contribution of sex and other bias-related variables, aligning with clinical evidence. The fairness-aware models exhibited satisfactory discriminatory performances, comparable with the fairness-unaware models. More importantly, FAIM significantly mitigated the biases measured by fairness metrics by 53.5%-57.6% for the MIMIC-IV-ED case, and 17.7%-21.7% for the SGH-ED case, outperforming commonly used bias-mitigation methods.

**Discussion and Conclusions**

The framework of FAIM can generate a model that significantly improves fairness, while preserving the model performance achieved by the performance-optimized yet fairness-unaware model. The interactive process of yielding a fairness-aware model facilitates domain experts to participate in the modeling process, allowing for enhanced multidisciplinary collaboration in shaping contextualized AI fairness.

# 1. Introduction

As artificial intelligence (AI) gains prominence in high-stakes fields like healthcare, concerns about fairness have grown.[1-3] Biases (as opposed to fairness) in machine learning arise when "sensitive" factors (e.g., age, sex, gender, race, ethnicity, or socio-economic status, etc.) unjustly skew decision-making.[4,5] In healthcare, a biased model can unjustly influence life-altering decisions, such as disease diagnosis[6,7] and organ allocation[8]. Clinical AI fairness is challenging due to healthcare's complexity and the impact of social determinants, and its integration into clinical decision-making is crucial to prevent the escalation of health disparities[9-11].

A seemingly straightforward approach to reduce bias is to simply exclude sensitive variables from the decision-making process, also named "Under blindness", but it has been deemed undesirable, particularly in cases where subpopulations (such as by sex) are distinct.[12,13] Numerous methods have been developed to systematically mitigate biases, categorized by the stage of the modeling process at which they operate.[14,15] Bias mitigation can occur pre-process through data adjustment (e.g., sampling[16] and reweighing[16]), in-process via direct fair model development (e.g., regularization[17-19] and representation learning[20,21]), or post-process by altering model outputs (e.g., equalized odds post-process[22]). However, pre-process methods often encounter challenges when addressing biases that involve intersecting multiple attributes (e.g., race and sex)[23,24]. Post-process approaches, by modifying outputs, fail to address the root causes of biases, leaving predictions altered but not clarified.[25]

Fairness in machine learning is hindered by a lack of interpretability, especially those employing models with black-box architectures for bias mitigation. Interpretability concerns also arise when randomization is utilized to post-process individual predictions (e.g., changing a positive one into negative) for the purpose of group fairness, without clear clinical justification[22]. In healthcare, the contextual nature of fairness necessitates specialized knowledge of clinicians to ensure that fairness definitions align with clinical realities.[11,12] However, the lack of interpretability poses formidable obstacles to their active participation in the modeling process, challenging the mutual understanding between AI developers and clinicians.

Another common limitation of existing bias-mitigation methods is that they often compromise model performance, as measured by machine learning metrics like area under the curve (AUC). This drawback can hamper practical application, cast doubt on the claimed fairness due to increased uncertainty, and potentially lead to severe and unforeseen consequences. According to a recent empirical study[26], approximately 50% of current bias-mitigation methods degraded model performance, whereas in 25% cases, they worsened both fairness and performance. Specifically, Pfohl et al.[27] empirically demonstrated that the methods penalizing prediction discrepancies (which are often interpreted as evidence of bias) can almost universally decrease multiple aspects of performance. Nevertheless, it has been empirically shown that such a trade-off between model performance and fairness is not inevitable.[5]

To bridge the gap in clinical AI fairness, we propose a new fairness-aware interpretable modeling (FAIM) framework (Figure 1) to achieve fairness with enhanced interpretability without sacrificing model performance. FAIM operates within a set of nearly-optimal models that offer high performance without necessarily reaching optimality[28]. By leveraging the varying degrees of model reliance on variables (including sensitive variables) within the cloud,[29,30] FAIM can identify alternative model formulations that improve model fairness without significantly impairing performance. Additionally, FAIM also examines the impact of excluding some (or all) sensitive variables on model fairness, visualizes the findings to help clinicians select a fairness-enhanced model with reasonable interpretation and contextualization, and employs SHapley Additive exPlanations (SHAP)[31] to clarify variable importance changes due to the fairness enhancement to further improve interpretation. We illustrate our method in the prediction of hospital admission in the emergency department using two large-scale clinical datasets, focusing on reducing potential bias due to two sensitive variables, i.e., race and sex. Although we use clinical case studies, the framework introduced has broader applicability across multiple domains and diverse tasks.

## 2. Results

We implemented FAIM on two large clinical datasets to predict hospital admission: Medical Information Mart for Intensive Care IV Emergency Department (MIMIC-IV-ED)[32] and data collected from the emergency department of Singapore General Hospital (SGH-ED)[33], with characteristics summarized in eTable 1-2. We aimed to mitigate biases related to sex and race, using logistic regression as the fairness-unaware baseline model ("baseline" for short). As demonstrated in the two datasets, the FAIM framework transparently navigated the exclusion of sensitive variables, and the FAIM output models (i.e., fairness-aware models, "FAIM" for short) had satisfactory discriminatory performance and significantly improved model fairness compared to other commonly used bias-mitigation methods.

### 2.1 Model selection among nearly-optimal models

After generating a set of nearly-optimal models (see details in subsection "nearly-optimal model generation" in Methods), we located the fairness-aware model via the interactive interface of FAIM. For the SGH-ED data, Figure 2 showcases the fairness distribution of these nearly-optimal models, aiding in model selection with data-driven evidence. In Figure 2A, regarding individual fairness metrics (equalized odds, equal opportunity and balanced error rate [BER], as defined in Table 1), these nearly-optimal models displayed diverse fairness profiles. Across panels, these models were jointly assessed using the proposed Fairness Ranking Index (FRI), which aggregates these individual fairness metrics for a comprehensive assessment of fairness (see Methods for detailed definition). As visualized in Figure 2B and quantified in Figure 2C, among top-10 models, the top-5 models excluded race, whereas the sixth-ranked model excluded both race and sex (shown in the golden box). Conversely, none of the models in the bottom-100 excluded race (shown in the grey box). We selected the top-1 model (i.e., model ID 224) as the FAIM output, which included the sex variable. This decision was informed by the similar distribution of nearly-optimal models regardless of sex variable exclusion, indicating minimal bias caused by sex (Figure 2C), coupled with the absence of clinically detected sex biases in Singapore's hospital admissions so far.

For the MIMIC-IV-ED dataset, FAIM's default choice was the top-1 model (i.e., model ID 681) with both race and sex excluded. This preference aligned with nine out of the top-10 models, as depicted in eFigure 1. Similarly, we retained the top-1 model as the FAIM output model, reflecting the majority preference of the top-ranking models and addressing known racial and sex disparities in emergency department care outcomes in the United States[34-36].

## 2.2 Evaluation of model fairness

With the FAIM models above, we evaluated model fairness on the split-out test set. FAIM consistently outperformed both baseline models and other bias-mitigation methods in model fairness, except for the method Reductions (see Table 2 for MIMIC-IV-ED data and Table 3 for SGH-ED data). FAIM's improvement in fairness was statistically significant ($p<0.001$), yielding a 53.5%-57.6% enhancement in fairness metrics for the MIMIC-IV-ED case, and a 17.7%-21.7% enhancement for the SGH-ED case. Importantly, while the Reductions method did improve fairness, it came at the cost of prediction performance, which led to impartially poor performance across different subgroups (more detail on performance evaluation in Section 3.2).

Our ablation study suggests that FAIM's advantages extend well beyond merely omitting sensitive variables, referred to as the method "under blindness". Indeed, regarding fairness metrics, FAIM's performance improvement over "under blindness" was notable—ranging from 6.62%-9.5% for the MIMIC-IV-ED dataset (Table 2) and 6.69%-7.09% for the SGH-ED dataset (Table 3).

Beyond fairness metrics, we conducted subgroup analyses for sensitive variables to evaluate FAIM's impact on fairness. Table 4 shows that FAIM substantially reduced disparities in both true positive and true negative rates across racial subgroups for both datasets. For sex subgroups, FAIM notably reduced the disparities for the MIMIC-IV-ED dataset, while its effect on the SGH-ED dataset was not consistently significant, given the relatively small initial disparities present in the baseline model.

To further explore the changes in model reliance on the variables, we conducted SHAP analysis to compare the fairness-unaware (i.e., baseline) and fairness-aware (i.e., FAIM) models. Beyond the sensitive variables of sex and race, for both datasets, the FAIM models preserved the variable contributions of less-sensitive variables similar to the baseline models, while making minor changes to address fairness. As indicated by the SHAP explanations (Figure 3), the FAIM models preserved the triage score Emergency Severity Index (ESI) for MIMIC-IV-ED and Patient Acuity Category Scale (PACS) for Singapore, but downplayed the "pain scale" in the MIMIC-IV-ED case (Figure 3A).

## 2.3 Evaluation of model performance and statistical functionality

With improved model fairness, we then evaluated FAIM from the perspective of model performance. Our analysis of classification metrics—AUC, sensitivity, and specificity, detailed in Table 2-3—revealed that FAIM maintained comparable performance to the baseline models. For the MIMIC-IV-ED dataset, FAIM achieved an AUC of 0.786 (95% CI: 0.783-0.789), closely aligned with the baseline model's AUC of 0.790 (0.787-0.793). Similarly, in the SGH-ED dataset, FAIM's AUC was 0.802 (0.801-0.804), which nearly matched the baseline model's AUC of 0.804 (0.802-0.806).

The sensitivity and specificity values achieved by FAIM were also on par with the baseline models. In MIMIC-IV-ED data, FAIM achieved a sensitivity of 0.725 (0.720-0.729) and specificity of 0.711 (0.707-0.715), compared to the baseline model with a sensitivity of 0.715 (0.711-0.720) and specificity of 0.724 (0.720-0.728). In SGH-ED data, FAIM delivered a sensitivity of 0.719 (0.717-0.721) and specificity of 0.746 (0.744-0.748), while the baseline model had a sensitivity of 0.713 (0.710-0.715) and specificity of 0.753 (0.751-0.755). Among other bias mitigation methods, in-process methods such as Reductions and post-process methods such as equalized odds post-process heavily degraded the classification performance.

FAIM models' odds ratios as well as the corresponding statistical significance were closely aligned with the fairness-unaware models, with minor changes addressing fairness (Figure 4)—unlike some conventional bias-mitigation methods that directly sacrifice such statistical functionality. The sensitive variables race and/or sex, which were statistically significant in the baseline models, were automatically excluded owing to their minimal effects on model predictive ability. Specifically, in the case of MIMIC-IV-ED data, the "pain scale" variable became less significant in FAIM (Figure 4), aligning with its reduced variable importance as shown in SHAP analyses (Figure 3).

## 3. Discussion

The pursuit of AI fairness is of increasing importance, particularly in high-stakes decision-making[1-3]. We contribute to this field by introducing FAIM, a fairness-aware, interpretable framework, to build well-performing yet fair models. Our findings reveal that fairness can be enhanced without sacrificing model performance. By emphasizing transparency and interpretability, FAIM could also promote active clinician engagement and foster multi-disciplinary collaboration.

## 3.1 Enhancing model fairness without impairing model performance

Beyond merely enhancing fairness, it is crucial for bias-mitigation methods to uphold satisfactory performance (e.g., AUC) and statistical functionality (e.g., odds ratio). A model's performance is paramount; without it, the model's utility in clinical decision-making is compromised[37]. Some fairness methods, particularly post-process methods, often sacrifice model performance[38] and raise concerns about their practicality. Our evaluations also reveal that these methods tend to reduce overall AUCs without reliably diminishing disparities (for example, equalized odds post-process, as shown in Table 2 and Table 3). Statistical functionality, for example, odds ratios and their confidence intervals, is also essential for elucidating model outputs and informing decision-making. Hence, it may be insufficient to merely provide binary predictions without details about why a model made a certain prediction or under what conditions the predictions hold, as is the case with most conventional bias-mitigation methods. In the demonstration with the logistic regression model on both MIMIC-IV-ED and SGH-ED datasets, FAIM could output odds ratios and confidence intervals for most less-sensitive variables, aligning with baseline models (Figure 3). These findings indicate that FAIM can provide a viable, fairness-aware alternative to the original machine learning models that are optimized for performance.

## 3.2 The importance of interpretability in fairness modeling

The interpretability in FAIM is multi-layered. Firstly, its underlying principle is intuitive which is to prioritize fairer models among a set of nearly-optimal models to improve fairness. Secondly, the process of model selection is transparent and interactive, as shown in Figure 2 (described in the subsection "Model selection with contextualized fairness" in Methods). Such an interactive process can assist in integrating data-driven evidence and clinical expertise, especially regarding sensitive variables that require contextualized consideration rather than a one-fits-all approach. Moreover, the SHAP analyses can illustrate changes in variable contribution between fairness-unaware and fairness-aware models, for example, the diminished importance of variable "pain description" on MIMIC-IV-ED data.

While many methods may produce bias-mitigated predictions, they may not always fundamentally address algorithmic bias or elucidate the modifications made to enhance fairness[25]. For example, the post-process method FaiRS developed by Coston et al[39], is also grounded in "near-optimality". FaiRS thoughtfully adjusts binary predictions to promote group fairness, occasionally by altering a negative prediction to a positive one. FAIM further underscores the role of model interpretability in improving fairness and showcases that model interpretability can serve as a tool to disclose model information, verify the effects of bias mitigation strategies, and potentially enhance trust.

### 3.3 Keeping domain experts in the loop

Clinical evidence plays a crucial role in reducing model biases, especially when dealing with sensitive variables in the decision-making process. The results of FAIM in the MIMIC-IV-ED case resonate with literature that highlights racial and sex disparities in emergency department care outcomes in the United States[34-36]. Beyond excluding sensitive variables, FAIM also attenuated the variable of pain description, which may carry inherent sex biases[40,41].

Model fairness can have geographical variations.[23] As detailed in Table 4, the original models displayed race and sex disparities which were more pronounced in the US dataset compared to the Singapore dataset, and FAIM also showed varying effects in addressing these disparities. Additionally, FAIM addresses disparities across both race and sex in the US data while primarily race in the Singapore data.

Advancing AI fairness in healthcare necessitates domain expertise to steer the process of bias mitigation effectively. FAIM's user-friendly graphical interface enables clinicians to directly influence model development with their insights. It also helps the validation of clinical judgment with data-driven evidence and encourages dialogue in the presence of divergences. These insights on fairness notions, critical metrics, and clinical evidence related to sensitive variables can ultimately shape a more equitable and contextualized approach to patient care.

### 3.4 The scalability of FAIM and future works

FAIM advocates that bias mitigation should occur concurrently with the modeling process, instead of a one-time step, given that all stages can inject bias into the final decision-making.[11] Despite FAIM being an in-process type method, it can synergize with pre-process strategies like Reweigh to tackle data underrepresentation, which is a known source of data bias, particularly when subgroup patterns are pronounced[38]. Additionally, methods that uncover data biases and abnormalities[42,43] can be integrated into the framework of FAIM for data pre-processing. In addition to its flexibility in module plug-in, the framework of FAIM also offers adaptability to other machine learning models (e.g., neural networks) in the future.

Although our study focused on separation-based metrics like equalized odds, equal opportunity, and equality of BER, the FAIM framework can also integrate other widely-used fairness definitions, for example, independence-based metrics[44] such as statistical parity into the FRI composition. Independence-based metrics are useful when the outcome of interest lacks reliable ground truth, rendering separation-based metrics unreliable.[45] However, caution is advised in using independence-based metrics, given the risk of misapplication particularly in complex contexts of healthcare where biological differences and social biases often intertwine with each other.[11,45,46]

While FAIM was designed with a focus on clinical AI fairness, its versatility extends to other high-stakes domains, such as finance and criminal justice. Notably, achieving AI fairness can present inherently distinct challenges across fields. The involvement of domain experts in the modeling process is indispensable, as their deep expertise can ensure that models are grounded in the nuanced realities of each domain. This further emphasizes the crucial role of keeping humans in the loop to navigate the complexities effectively.

### 3.5 Limitations

This study has several limitations. Firstly, we only demonstrated our method in two datasets related to emergency medicine. Further validation is needed in a wide spectrum of clinical applications. Secondly, the procedure of pinpointing nearly-optimal models may benefit from further optimization. Moreover, there is room for improvement in the manner of obtaining confidence intervals for the alternative odds ratios to more closely align with the functionality of the original models.

## 4. Conclusion

FAIM is an interpretable and interactive framework capable of generating fairness-aware models. These models yielded by FAIM can achieve a significant improvement in fairness, without compromising the performance and functionality. The interpretability of this approach invites domain experts into the heart of the modeling process, fostering a richer multidisciplinary collaboration. This collaboration is crucial to crafting AI fairness that is tailored to specific contexts.

## 5. Methods

The FAIM framework consists of three modules to locate a "fairer" model in a set of nearly-optimal models: nearly-optimal model generation, model selection with contextualized fairness, and model explanation, as visualized in Figure 1. Let Y denote the outcome and $X_S = (X_{S_1}, \ldots, X_{S_k})$ denote a set of $k$ sensitive variables (e.g., sex and race), $X_{S'}$ represents a feature subset of sensitive variables in $X_S$, and $X_U = (X_{U_1}, \ldots, X_{U_p})$ collectively denote $p$ less-sensitive variables. Notably, $X_U$ are variables that do not directly contain sensitive information that may cause bias, but as shown in the clinical examples, some seemingly non-sensitive variables may indirectly link to sensitive information and be affected after fairness enhancement. A model built with variables $X_U$ and $X_{S'}$ is denoted by $f(X_U, X_{S'})$. The loss function of this model is denoted as $L(f(X_U, X_{S'}), Y)$, along with the expected loss $E[L(f(X_U, X_{S'}), Y)]$.

## 5.1 Nearly-optimal model generation: composing integral Rashomon sets with multiple sensitive variables

Conventional efforts to develop an optimal model typically focus on model performance, which is to fit the data and minimize the expected loss functions. Nevertheless, as one approaches the theoretically optimal model, many models can fit the data with comparable loss, forming a set of nearly-optimal models—an intriguing phenomenon known as the Rashomon effect in statistics.[28,29] Within this Rashomon set of nearly-optimal models, models can differ in their reliance on covariate information.[30,47] This diversity leads to variations in fairness profiles, especially concerning sensitive variables.

Let $f^*_{(U,S')}$ denote the optimal model that minimizes the expected loss in the model family $F_{(U,S')}$ that is built with less-sensitive variables $X_U$ and sensitive variables $X_{S'}$. The $S'$-specific Rashomon set is defined as:

$$R_{(U,S')}\left(\epsilon_0, f^*_{(U,S')}, F_{(U,S')}\right) = \left\{f \in F_{(U,S')} \mid E[L(f, Y)] \leq (1+\epsilon_0) E\left[L\left(f^*_{(U,S')}, Y\right)\right]\right\},$$

where "near-optimality" is controlled by the small factor $\epsilon_0 > 0$. Particularly, for parametric models that can be fully represented by its coefficients $\beta_{(U,S')}$, e.g., the family of logistic regression, the Rashomon set can be converted into a set of coefficients:

$$R_{(U,S')}\left(\epsilon_0, \beta^*_{(U,S')}, B_{(U,S')}\right) = \left\{\beta \in B_{(U,S')} \mid E[L(f_\beta, Y)] \leq (1+\epsilon_0) E\left[L\left(f^*_{(U,S')}, Y\right)\right]\right\},$$

where $B_{(U,S')}$ is the coefficients space of models in $F_{(U,S')}$, and $\beta^*_{(U,S')}$ is the coefficients of the optimal model $f^*_{(U,S')}$.

To objectively evaluate the effects of sensitive variables on model performance and fairness, we consider different cases of variable selection for $X_{S'}$. The cases include no exclusion (i.e., $X_{(U,S)}$, the baseline case), complete exclusion (i.e., $X_U$, "Under blindness"), and all possible cases of partial exclusion. We defined the Integral Rashomon Set (IRS) with all the cases of $S'$ considered that can be expressed as:

$$R\left(\epsilon, \beta^*_{(U,\cdot)}, B_{(U,\cdot)}\right) = \bigcup_{\substack{S' \subseteq S \\ \beta^*_{(U,S')} \in R_{(U,S)}\left(\epsilon_0, \beta^*_{(U,S)}, B_{(U,S)}\right)}} R_{(U,S')}\left(\epsilon_0, \beta^*_{(U,S')}, B_{(U,S')}\right),$$

where $\epsilon = (\epsilon_0 + 1)^2 - 1$. The case-specific near-optimality determined by $\epsilon_0$ should be more stringent to guarantee the overall near-optimality of IRS determined by $\epsilon$, i.e., $\epsilon > \epsilon_0 > 0$ (see more details in Supplementary eMethod 1). In previous studies, $\epsilon$ is often set at 5%.[30,47]

We employ the method of rejection sampling[30,48] to identify the nearly-optimal models, that is, to generate random samples of the coefficient vector and reject those with corresponding expected loss out of "near-optimality". Specifically, to fully represent the case-specific Rashomon set $R_{(U,S')}\left(\epsilon_0, \beta^*_{(U,S')}, B_{(U,S')}\right)$, the $i$-th sample of coefficient vector is generated from a multivariable normal distribution $N\left(\beta^*_{(U,S')}, k_i \Sigma^*_{(U,S')}\right)$ where $\beta^*_{(U,S')}$ and $\Sigma^*_{U,S'}$ are the coefficient vector and variance-covariance matrix of the optimal model $f^*_{(U,S')}$ based on less-sensitive variables $X_U$ and sensitive variables $X_{S'}$. The control parameter of scope-width $k_i$ is drawn from a uniform distribution $U(u_1, u_2)$ with tunable parameters $u_1$ and $u_2$ to adjust the scope of sampling.

## 5.2 Model selection with contextualized fairness

The FAIM framework prioritizes the fairness notion of ensuring equality of model performance across subgroups (also named "separation-based" fairness)[44,49] This fairness notion encompasses various fairness metrics that essentially measure the gaps of performance (e.g., accuracy, AUC, sensitivity, specificity, etc.) among subgroups, where smaller gaps indicating better fairness[49]. For example, equal opportunity[22] emphasizes equal sensitivity (i.e., true positive rate) across subgroups, defined as the maximal discrepancy in sensitivity values across subgroups. To comprehensively rank the nearly-optimal models in the IRS i.e., $R\left(\epsilon, \beta^*_{(U,\cdot)}, B_{(U,\cdot)}\right)$ and select a fairer one, we developed the Fairness Ranking Index (FRI). Inspired by the radar chart for comparing items across multiple dimensions, FRI is a holistic ranking measure that considers not only individual dimensions of fairness metrics $(m_1, m_2, \ldots, m_J)$ but also their interdependencies, calculated as:

$$FRI(f_\beta) = \frac{1}{\sum_{j=1}^{J} m_j(f_\beta) m_{j+1}(f_\beta)},$$

where $m_{J+1} := m_1$ to simplify notations. The highest-ranked model within $R\left(\epsilon, \beta^*_{(U,\cdot)}, B_{(U,\cdot)}\right)$ is chosen as the default fairness-aware model:

$$\tilde{\beta} = \underset{\beta \in R\left(\epsilon, \beta^*_{(U,\cdot)}, B_{(U,\cdot)}\right)}{\arg\max} FRI(f_\beta),$$

with a corresponding subset of sensitive variable(s) $X_{\tilde{S}}$. In this paper, we focus on well-established fairness metrics including equalized odds, equal opportunity, and BER, as defined in Table 1.

The process of model selection is designed to be interpretable and interactive. To facilitate such human-involved investigation, FAIM provides an interactive graphical interface (https://github.com/nliulab/FAIM) to display detailed information on each model upon hovering. As illustrated in Figure 2 based on SGH-ED data, Figure 2A visualized individual fairness metrics in each panel for the sampled nearly-optimal models, with joint fairness ranked by the FRI. The top-1 model highlighted in gold, is the default choice for the fairness-aware model. Nevertheless, users can assess alternative models with high rankings, such as the top-10 models, through a clinical lens, based on detailed data-driven evidence provided in Figure 2B-C. These high-ranked models may differ in how they handle sensitive variables yet have comparable model performance and fairness status. Clinical justification can jump in to support the final exclusion of sensitive variables. Therefore, the final fairness-aware model yielded by FAIM integrates domain knowledge with data-driven evidence to enhance model fairness without impairing performance.

As depicted in Figure 1, we adopted SHapley Additive exPlanations (SHAP)[31] values for interpretation to investigate how variable importance shifts before and after fairness enhancement. The fairness-aware model $f_{\tilde{\beta}}$ shares a similar model architecture with the fairness-unaware $f_{\beta^*_{(U,S)}}$, except it may exclude certain sensitive variables that do not contribute to model performance and degraded fairness. The vector $\tilde{\beta}$ serves as a nearly-optimal solution that minimally deviates from $\beta^*_{(U,S)}$ to execute fairness adjustments. In addition, $f_{\tilde{\beta}}$ holds the same functionality with $f^*_{(U,S)}$ in implementation. Take logistic regression for example, the coefficient $\tilde{\beta}$ can still be interpreted as the vector of log values of odds ratios, corresponding to the model $f_{\tilde{\beta}} = \frac{1}{1+e^{-\tilde{\beta} X_{(U,\tilde{S})}}}$ and logit loss $L(f_{\tilde{\beta}}, Y)$. The standard errors for $\tilde{\beta}$ are estimated using Fisher's information $I(\tilde{\beta})$, mirroring the method used for $\beta^*_{(U,S)}$ based on $I(\beta^*_{(U,S)})$.

## 5.3 Data and study design

To illustrate the clinical application of the FAIM framework, we used two datasets from different populations. The first dataset was derived from the Medical Information Mart for Intensive Care IV Emergency Department (MIMIC-IV-ED)[32], a publicly available database of emergency department admissions at the Beth Israel Deaconess Medical Center (BIDMC) in the United States between 2011 and 2019. The second dataset, referred to as the Singapore General Hospital Emergency Department (SGH-ED) database[33], was acquired from a tertiary hospital in Singapore, with data of approximately 1.8 million ED visits between 2008 and 2020.

Our predictive models for hospital admission ($Y$), informed by a previous study[50], utilized predictors including demographic data, vital signs, triage score, and health records (see more details in supplementary eTable 3). For MIMIC-IV-ED data, we excluded patients with age below 18[50], while for SGH-ED data, we excluded patients who were not Singapore residents or were aged below 21.[47] In MIMIC-IV-ED data, we recategorized race into five groups—Asian, Black, Hispanic, White and others, and binarized Emergency Severity Index (ESI) into low risk (3-5) and high risk (1-2)[51]. In SGH-ED data, we recategorized race into four groups: Chinese, Indian, Malay and others, and recategorized the triage score—Patient Acuity Category Scale (PACS) into three levels: P1, P2 and P3-4[52]. Given the interplay of racism and sexism[23,24], we treated race and sex as sensitive variables in predicting hospital admissions.

Each dataset was randomly split into training (70%), validation (10%), and test (20%) sets. We respectively utilized trainining set for the generation of nearly-optimal models, validation set for model selection with contextualized fairness and parameter tuning, and test set for model evaluation. In addition, to illustrate variable importance using SHAP, we utilized subsamples of training and validation sets respectively as background and explanation data.

## 5.4 Statistical analysis

We compared FAIM with common bias-mitigation methods, including pre-process method Reweigh[16], in-process method Reductions[19], and post-process method equalized odds post-processing[22]. A conventionally trained logistic regression model served as the fairness-unaware baseline model. We evaluated model fairness with equalized odds, equal opportunity, and equality of BER. We assessed model performance using the area under the curve (AUC) values, sensitivity and specificity, with 95% confidence intervals (CIs) reported. The sensitivity and specificity values were yielded based on optimal thresholds determined by Youden's J statistics[53]. Comparisons between bias mitigation methods were performed using the t-test with bootstrap sampling. The data analysis and model building were performed using R version 4.0.2 (The R Foundation for Statistical Computing) and python version 3.9.7. A two-sided p<0.05 was considered statistically significant.

**Table 1.** Separation-based fairness metrics

| Description | Fairness metrics | Formulas |
|---|---|---|
| General description<br><br>The model have similar machine learning performance among subgroups, e.g., same false positive rates. The predictions can have relationship with the sensitive variables.<br><br>Example scenario<br><br>Disease detection in radiology in terms of race/ethnicity[54]<br><br>Statistical assumption<br>Conditional independence<br><br>($\hat{Y} \perp X_S \mid Y$) | Equalized odds | $\Delta_{EOD} = \max_Y \text{Range}_S(E[\hat{Y}|X_{(U,S)}, Y])$ |
| | Equalized opportunity | $\Delta_{EOP} = \text{Range}_S(E[\hat{Y}|X_{(U,S)}, Y = 1])$ |
| | Balanced error rate | $\Delta_{BER} = \text{Range}_S(P(\hat{Y} = 1|Y = 0, X_{(U,S)}) + P(\hat{Y} = 0|Y = 1, X_{(U,S)}))$ |

**Table 2.** Two-dimension evaluation (fairness and performance) for bias mitigation methods (MIMIC-IV-ED data)

|  | Fairness metrics[1] | | | Performance metrics | | |
|---|---|---|---|---|---|---|
|  | Equal Opportunity | Equalized Odds | BER equality[2] | AUC | Sensitivity[3] | Specificity[3] |
| Baseline[4] | 0.316 | 0.316 | 0.301 | 0.790 [0.787, 0.793] | 0.715 [0.711, 0.720] | 0.724 [0.720, 0.728] |
| FAIM | 0.134 | 0.147 | 0.140 | 0.786 [0.783, 0.789] | 0.725 [0.720, 0.729] | 0.711 [0.707, 0.715] |
| Reweigh (pre-process) | 0.206 | 0.206 | 0.183 | 0.789 [0.786, 0.792] | 0.709 [0.704, 0.713] | 0.731 [0.726, 0.735] |
| Reductions (in-process) | 0.037 | 0.037 | 0.032 | 0.706 [0.703, 0.709] | 0.657 [0.652, 0.662] | 0.755 [0.751, 0.759] |
| Equalized odds post-processing (post-process) | 0.734 | 0.734 | 0.554 | 0.594 [0.591, 0.596] | 0.335 [0.330, 0.340] | 0.852 [0.849, 0.855] |
| Under blindness[5] | 0.159 | 0.159 | 0.155 | 0.787 [0.784, 0.790] | 0.707 [0.703, 0.711] | 0.729 [0.724, 0.733] |

[1] Smaller values indicate higher levels of fairness

[2] BER: equality of the combination of true positive rate and true negative rate

[3] The thresholds were determined by Youden's J index for methods that can yield predictive probabilities (i.e., original logistics regression, "FAIM", "Reweigh", "Under blindness"). In-process method "Reductions" and post-process method "equalized-odds post-process" directly generated the binary prediction.

[4] Baseline: the original logistics regression model, i.e., fairness-unaware counterpart

[5] Under blindness: the logistics regression with sensitive variables excluded

**Table 3.** Two-dimension evaluation (fairness and performance) for bias mitigation methods (SGH-ED data)

|  | Fairness metrics[1] | | | Performance metrics | | |
| --- | --- | --- | --- | --- | --- | --- |
|  | Equal Opportunity | Equalized Odds | BER equality[2] | AUC | Sensitivity[3] | Specificity[3] |
| Baseline[4] | 0.299 | 0.299 | 0.237 | 0.804 [0.803, 0.806] | 0.713 [0.710, 0.715] | 0.753 [0.751, 0.755] |
| FAIM | 0.234 | 0.234 | 0.195 | 0.802 [0.801, 0.804] | 0.719 [0.717, 0.721] | 0.746 [0.744, 0.748] |
| Reweigh (pre-process) | 0.244 | 0.244 | 0.212 | 0.803 [0.802, 0.805] | 0.712 [0.710, 0.715] | 0.750 [0.748, 0.752] |
| Reductions (in-process) | 0.035 | 0.036 | 0.036 | 0.664 [0.662, 0.665] | 0.542 [0.540, 0.545] | 0.785 [0.784, 0.787] |
| Equalized odds post-processing (post-process) | 0.685 | 0.685 | 0.490 | 0.601 [0.600, 0.603] | 0.327 [0.323, 0.330] | 0.875 [0.874, 0.876] |
| Under blindness[5] | 0.252 | 0.252 | 0.209 | 0.803 [0.802, 0.805] | 0.720 [0.717, 0.722] | 0.744 [0.742, 0.746] |

[1] Smaller values indicate higher levels of fairness

[2] BER equality: equality of the combination of true positive rate and true negative rate

[3] The thresholds were determined by Youden's J index for methods that can yield predictive probabilities (i.e., original logistics regression, "FAIM", "Reweigh", "Under blindness"). In-process method "Reductions" and post-process method "equalized-odds post-process" directly generated the binary prediction.

[4] Baseline: the original logistics regression model, i.e., fairness-unaware counterpart

[5] Under blindness: the logistics regression with sensitive variables excluded

**Table 4.** Subgroup analysis regarding sex and race

|      |              | MIMIC-IV-ED |                              |       | SGH-ED   |                  |       |
|------|--------------|-------------|------------------------------|-------|----------|------------------|-------|
|      |              | Baseline    | Under blindness[1]           | FAIM  | Baseline | Under blindness  | FAIM  |
| Race | $\Delta TPR$[2] | 0.251   | 0.133                        | **0.122** | 0.264 | 0.223          | **0.211** |
|      | $\Delta TNR$[3] | 0.224   | 0.130                        | **0.129** | 0.170 | 0.149          | **0.134** |
| Sex  | $\Delta TPR$ | 0.074       | 0.025                        | **0.018** | 0.015 | **0.001**      | 0.005 |
|      | $\Delta TNR$ | 0.086       | 0.034                        | **0.027** | 0.008 | **0.025**      | 0.027 |

[1] Under blindness: the logistics regression with sensitive variables excluded

[2] $\Delta TPR$: the gap of true positive rate among race/ethnicity or sex subgroups

[3] $\Delta TNR$: the gap of true negative rate among race/ethnicity or sex subgroups

**Figure 1**. The general framework of FAIM

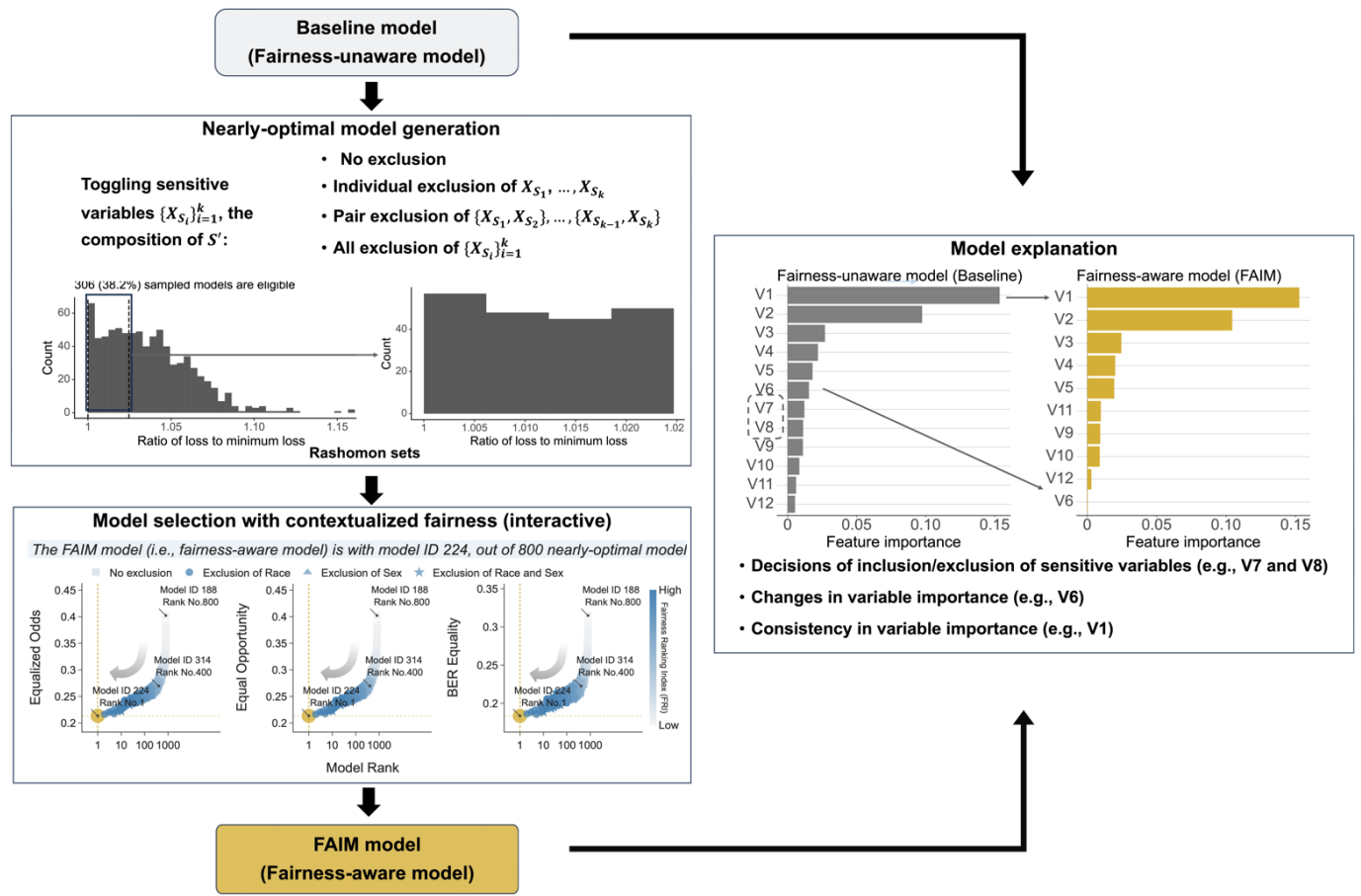

The FAIM framework consists of three modules to locate a "fairer" model in a set of nearly-optimal models: nearly-optimal model generation, model selection with contextualized fairness and model explanation.

**Figure 2.** The interactive plot for model selection, based on the nearly-optimal models' fairness evaluated on the validation set of SGH-ED data.

**A.**

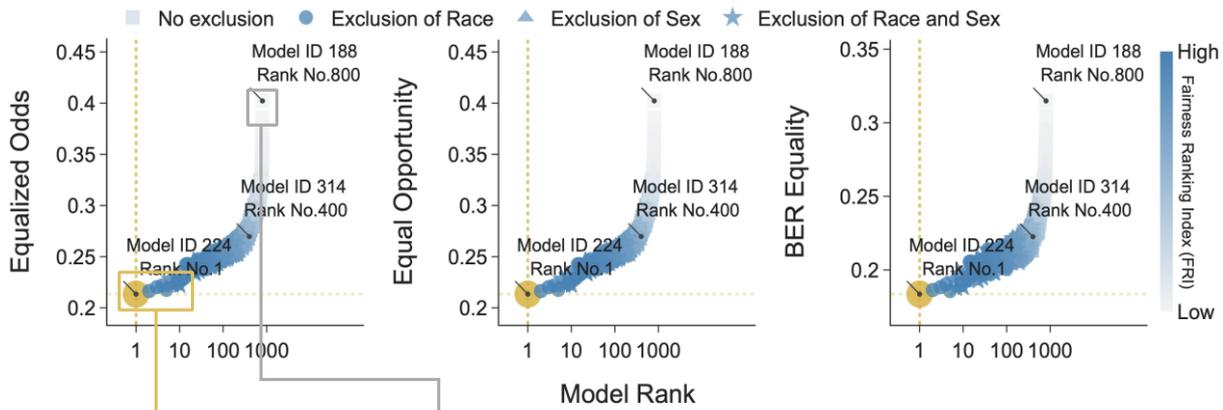

**B.**

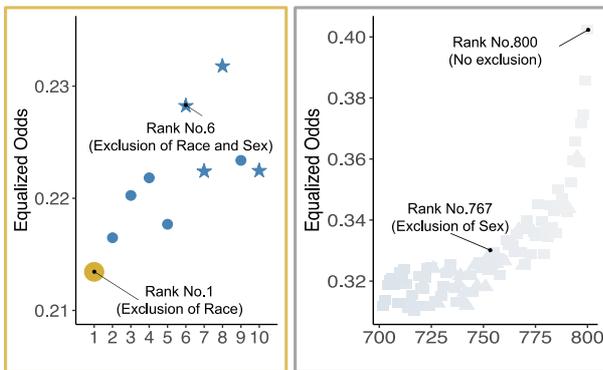

**C.**

Exclusion of Sensitive Variables on Fairness Rankings

| Exclusion cases | 1-10 ("most fair") | 11-700 | 701-800 ("least fair") | Highest ranking |
|---|---|---|---|---|
| No exclusion | 0 | 125 | 76 | No.289 |
| Exclusion of Race | 6 | 194 | 0 | No.1 |
| Exclusion of Sex | 0 | 176 | 24 | No.20 |
| Exclusion of Race and Sex | 4 | 196 | 0 | No.6 |

**A.** The graphical interface ranks nearly-optimal models based on fairness metrics—equalized odds, equal opportunity, and BER equality. Model 224 is ranked first by the Fairness Ranking Index (FRI), highlighted as the default fairness-aware model. Users can interactively engage with the data by hovering over points to display model details, zooming in on sections like the top-10 models, and adjusting panel views. **B.** Illustration of the panel of "equalized odds" metric with top-10 models showcased in the gold box, showing four models (stars) excluding both sex and race, and six models (circles) excluding only sex. The bottom-100 models, showcased in the grey box, predominantly include no exclusions (rectangles), with a minority excluding sex (upward triangles). **C.** A tabulation of models based on the exclusion cases of sensitive variable(s), detailing the counts of models in different fairness ranges—from "most fair" to "least fair". The right-most column records the highest ranking obtained by the nearly-optimal models for each exclusion case.

**Figure 3.** Variable importance analysis in both fairness-unaware model (i.e., "Baseline") and fairness-aware (i.e, "FAIM") model based on MIMIC-IV-ED (A) and SGH-ED datasets (B).

**A.**

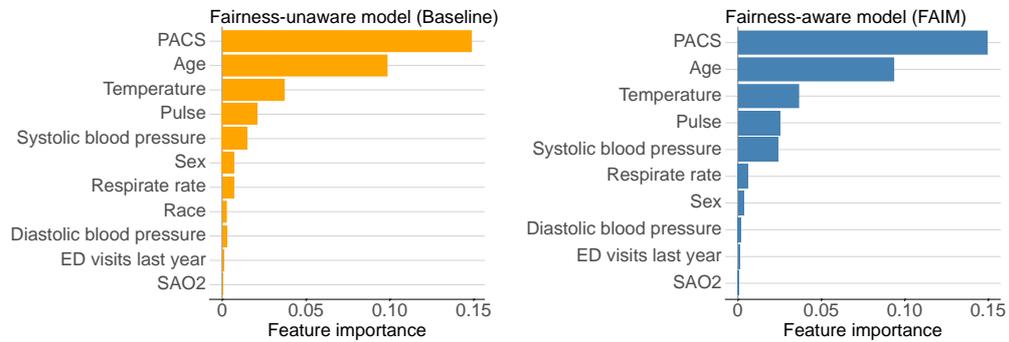

**B.**

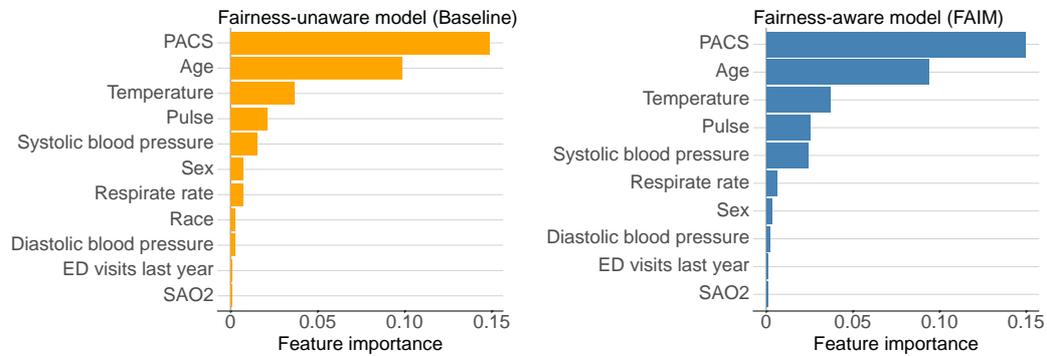

The fairness-aware model refers to the ranking No.1 fairest model yielded by FAIM, and the fairness-unaware model refers to the baseline model, i.e. the logistics regression model.

**Figure 4.** Comparison of the odds ratios between fairness-unaware ("Baseline") and fairness-aware ("FAIM") models based on MIMIC-IV-ED (A) and SGH-ED datasets (B).

**A.**

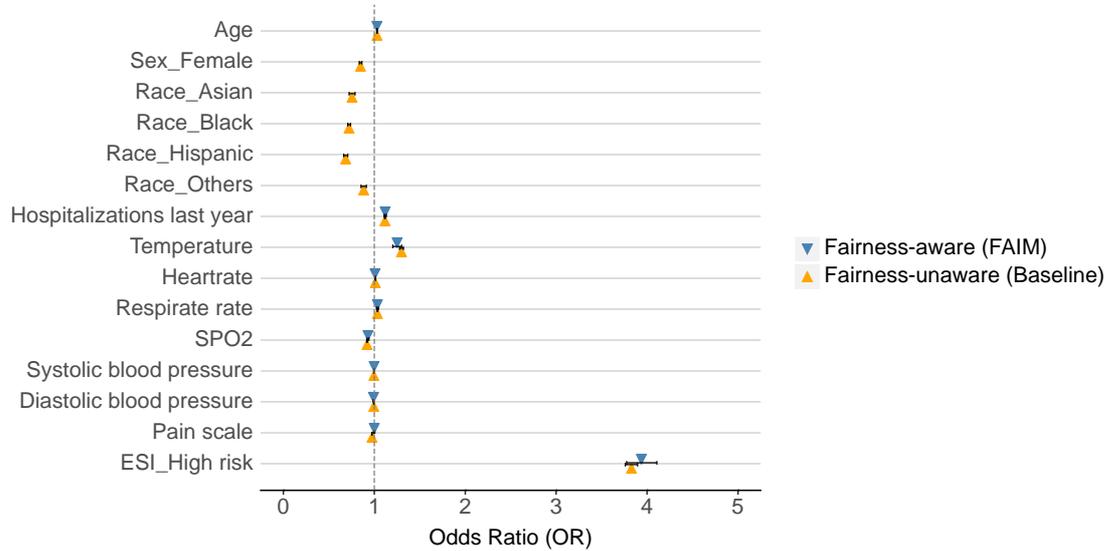

**B.**

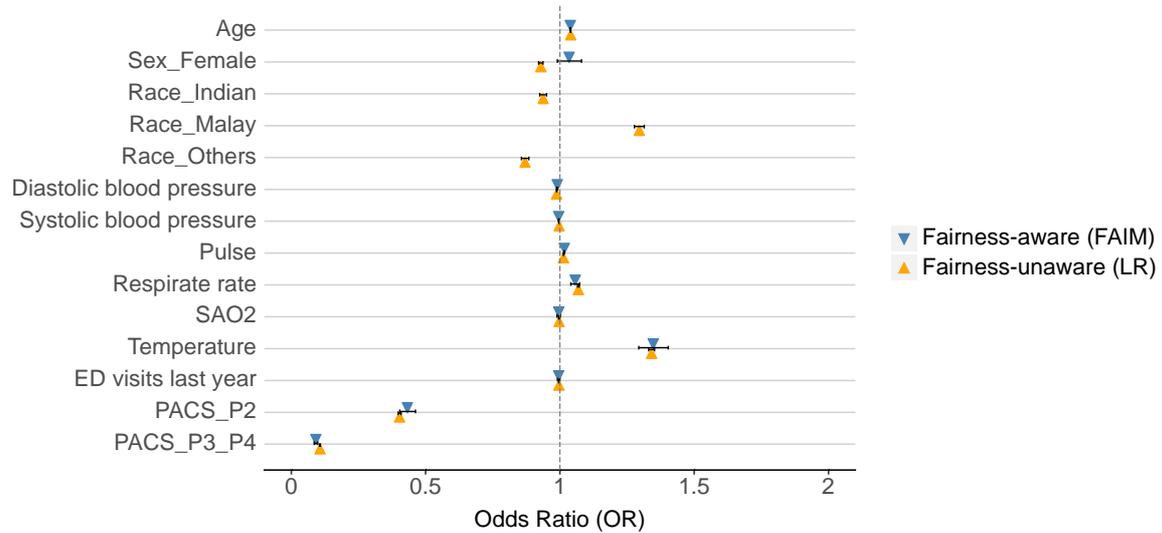

On the MIMIC-IV-ED dataset, the framework of FAIM excluded both sensitive variables race and sex from modeling to address fairness. As a result, the FAIM model did not have odds ratios for these two variables. Similarly, on the SGH-ED dataset, the variable race was excluded.

# Supplementary Materials

**eMethod 1. Mechanism of $\epsilon$ and $\epsilon_0$ for defining Rashomon set**

The Integral Rashomon Set (IRS), $R(\epsilon, \beta^*_{(U,\cdot)}, B_{(U,\cdot)})$, represents a "double near-optimal" subset of the actual Rashomon set, $R(\epsilon, \beta^*_{(U,S)}, B_{(U,S)})$. In other words, the exclusion case of feature subset $S/S'$ will be considered and the $S'$-specific cloud of "nearly optimal" models, $R_{(U,S')}(\epsilon_0, \beta^*_{(U,S')}, B_{(U,S')})$, will contribute to the composition of IRS, if and only if the $S'$-specific optimal model, $\beta^*_{(U,S')}$, is "nearly optimal" compared to the optimal modal, $\beta^*_{(U,S)}$, obtained in the original case:

If $\beta \in R(\epsilon, \beta^*_{(U,\cdot)}, B_{(U,\cdot)})$, then there exists a subset $S' \subseteq S$ such that $E[L(f_\beta, Y)] \leq (1+\epsilon_0) E\left[L\left(f^*_{(U,S')}, Y\right)\right]$ where $f^*_{(U,S')}$ is equivalent to $f_{\beta^*_{(U,S')}}$. Since $\beta^*_{(U,S')} \in R_{(U,S)}(\epsilon_0, \beta^*_{(U,S)}, B_{(U,S)})$ and $\epsilon = (\epsilon_0 + 1)^2 - 1$, we have $E\left[L\left(f^*_{(U,S')}, Y\right)\right] \leq (1+\epsilon_0) E[L(f^*_{(U,S)}, Y)]$. Thus,

$$\begin{aligned} E[L(f_\beta, Y)] &\leq (1+\epsilon_0) E\left[L\left(f^*_{(U,S')}, Y\right)\right] \\ &\leq (1+\epsilon_0)^2 \, E[L(f^*_{(U,S)}, Y)] \\ &= (1+\epsilon) E[L(f^*_{(U,S)}, Y)]. \end{aligned}$$

That is, $\beta \in R(\epsilon, \beta^*_{(U,S)}, B_{(U,S)})$.

**eTable 1.** Characteristics of MIMIC-IV-ED dataset

|  | Overall | Not admitted | Admitted | p-value |
|---|---|---|---|---|
| n | 418100 | 220276 | 197824 |  |
| **Age, mean (SD)** | 52.8 (20.6) | 46.3 (19.4) | 60.1 (19.5) | <0.001 |
| **Sex, n (%)** |  |  |  | <0.001 |
| Female | 227007 (54.3) | 126755 (57.5) | 100252 (50.7) |  |
| Male | 191093 (45.7) | 93521 (42.5) | 97572 (49.3) |  |
| **Race, n (%)** |  |  |  | <0.001 |
| Asian | 18321 (4.4) | 11197 (5.1) | 7124 (3.6) |  |
| Black | 92168 (22.0) | 55944 (25.4) | 36224 (18.3) |  |
| Hispanic | 34150 (8.2) | 22158 (10.1) | 11992 (6.1) |  |
| White | 242666 (58.0) | 113445 (51.5) | 129221 (65.3) |  |
| Others | 30795 (7.4) | 17532 (8.0) | 13263 (6.7) |  |
| **ESI, n (%)** |  |  |  | <0.001 |
| High risk[1-2] | 163430 (39.1) | 48103 (21.8) | 115327 (58.3) |  |
| Low risk[3-5] | 254670 (60.9) | 172173 (78.2) | 82497 (41.7) |  |
| **Systolic blood pressure, mean (SD)** | 134.9 (22.2) | 135.2 (20.7) | 134.5 (23.7) | <0.001 |
| **Heartrate, mean (SD)** | 85.0 (17.4) | 83.9 (16.3) | 86.3 (18.6) | <0.001 |
| **Diastolic blood pressure, mean (SD)** | 77.5 (14.7) | 78.8 (13.8) | 76.0 (15.6) | <0.001 |
| **Temperature, mean (SD)** | 36.7 (0.5) | 36.7 (0.5) | 36.7 (0.6) | <0.001 |
| **Pain scale, mean (SD)** | 4.2 (3.6) | 4.7 (3.6) | 3.6 (3.5) | <0.001 |
| **SpO$_2$, mean (SD)** | 98.4 (2.4) | 98.8 (2.0) | 97.9 (2.7) | <0.001 |
| **Respiratory rate, mean (SD)** | 17.6 (2.5) | 17.3 (2.1) | 17.9 (2.8) | <0.001 |
| **Hospitalizations last year, mean (SD)** | 1.0 (2.7) | 0.6 (2.2) | 1.4 (3.1) | <0.001 |

ESI: Emergency Severity Index; SpO$_2$: oxygen saturation as detected by the pulse oximeter.

**eTable 2.** Characteristics of SGH-ED dataset

|  | Overall | Not admitted | Admitted | p-value |
|---|---|---|---|---|
| n | 1716830 | 1074513 | 642317 | |
| **Age, mean (SD)** | 53.1 (19.3) | 47.3 (18.0) | 62.6 (17.4) | <0.001 |
| **Sex, median [Q1,Q3]** | | | | <0.001 |
| Female | 822284 (47.9) | 511060 (47.6) | 311224 (48.5) | |
| Male | 894546 (52.1) | 563453 (52.4) | 331093 (51.5) | |
| **Race, median [Q1,Q3]** | | | | <0.001 |
| Chinese | 1114566 (64.9) | 660744 (61.5) | 453822 (70.7) | |
| Indian | 229058 (13.3) | 158004 (14.7) | 71054 (11.1) | |
| Malay | 195334 (11.4) | 121211 (11.3) | 74123 (11.5) | |
| Other Races | 177872 (10.4) | 134554 (12.5) | 43318 (6.7) | |
| **PACS, n (%)** | | | | <0.001 |
| P1 | 202617 (11.8) | 54231 (5.0) | 148386 (23.1) | |
| P2 | 730445 (42.5) | 365453 (34.0) | 364992 (56.8) | |
| P3_P4 | 783768 (45.7) | 654829 (60.9) | 128939 (20.1) | |
| **Systolic blood pressure, mean (SD)** | 132.1 (22.9) | 131.2 (21.2) | 133.6 (25.5) | <0.001 |
| **Pulse, mean (SD)** | 81.4 (15.8) | 80.7 (14.4) | 82.6 (17.8) | <0.001 |
| **Diastolic blood pressure, mean (SD)** | 73.1 (12.8) | 74.0 (11.9) | 71.6 (14.1) | <0.001 |
| **Temperature, mean (SD)** | 36.6 (0.6) | 36.5 (0.6) | 36.6 (0.6) | <0.001 |
| **SaO$_2$, mean (SD)** | 98.2 (3.2) | 98.4 (3.0) | 97.9 (3.6) | <0.001 |
| **Respiratory rate, mean (SD)** | 17.6 (1.5) | 17.4 (1.3) | 17.8 (1.8) | <0.001 |
| **ED visits last year, mean (SD)** | 1.3 (4.8) | 1.1 (5.4) | 1.5 (3.5) | <0.001 |

PACS: Patient Acuity Category Scale; SaO$_2$: oxygen saturation of arterial blood.

**eTable 3.** Variable selection

|  | MIMIC-IV-ED data | SGH-ED data |
|---|---|---|
| Demographic data | age, race, sex | age, race, sex |
| Vitals | temperature, bp-systolic, bp-diastolic, heartrate, respirate rate, pain | temperature, bp-systolic, bp-diastolic, pulse, respirate rate |
| Triage score | ESI | PACS |
| Oxygenation | $SpO_2$ | $SaO_2$ |
| Health record | no. hospitalization in one year | no. emergency visit in one year |

ESI: Emergency Severity Index; PACS: Patient Acuity Category Scale; $SpO_2$: oxygen saturation as detected by the pulse oximeter; $SaO_2$: oxygen saturation of arterial blood.

**eFigure 1.** The interactive plot for model selection, based on the nearly-optimal models' fairness evaluated on the validation set (MIMIC-IV-ED).

**A.**

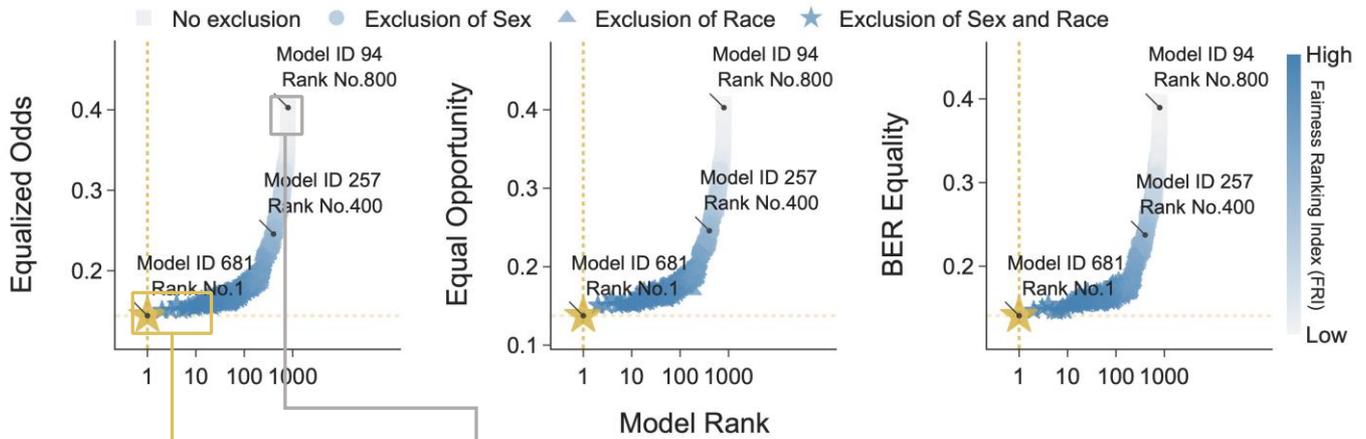

**B.**

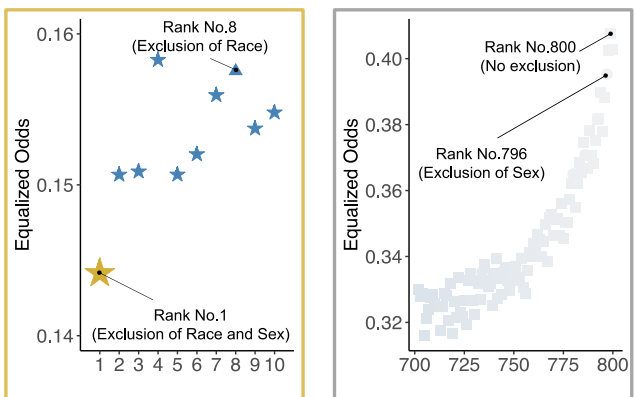

**C.**

Exclusion of Sensitive Variables on Fairness Rankings

| Exclusion cases | 1-10 ("most fair") | 11-700 | 701-800 ("least fair") | Highest ranking |
|---|---|---|---|---|
| No exclusion | 0 | 105 | 96 | No.426 |
| Exclusion of Race | 1 | 199 | 0 | No.8 |
| Exclusion of Sex | 0 | 196 | 4 | No.284 |
| Exclusion of Race and Sex | 9 | 190 | 0 | No.1 |

**A.** The graphical interface ranks nearly-optimal models based on fairness metrics—equalized odds, equal opportunity, and BER equality. Model 681 is ranked first by the Fairness Ranking Index (FRI), highlighted as the default fairness-aware model. Users can interactively engage with the data by hovering over points to display model details, zooming in on sections like the top-10 models, and adjusting panel views. **B.** Illustration of the panel of "equalized odds" metric with top-10 models showcased in the gold box, showing nine models (stars) excluding both sex and race, and one models (upwards triangle) excluding only race. The bottom-100 models, showcased in the grey box, predominantly include no exclusions (rectangles), with a minority excluding sex (upward triangles). **C.** A tabulation of models based on the exclusion cases of sensitive variable(s), detailing the counts of models in different fairness ranges—from "most fair" to "least fair". The right-most column records the highest ranking obtained by the nearly-optimal models for each exclusion case.